\documentclass[10pt,twocolumn,letterpaper]{article}

\PassOptionsToPackage{table}{xcolor}
\usepackage[pagenumbers]{cvpr}
\usepackage{multirow}
\usepackage{afterpage}

\usepackage{booktabs,array,tabularx,arydshln}

\definecolor{cvprblue}{rgb}{0.21,0.49,0.74}
\usepackage[breaklinks,colorlinks,allcolors=cvprblue,bookmarks=true,bookmarksnumbered=true]{hyperref}
\hypersetup{
  pdftitle={Bidirectional Cross-Modal Prompting for Event-Frame Asymmetric Stereo},
  pdfsubject={Computer Vision, Deep Learning, Robotics},
  pdfauthor={Ninghui Xu, Fabio Tosi, Lihui Wang, Jiawei Han, Luca Bartolomei, Zhiting Yao, Matteo Poggi, Stefano Mattoccia},
  pdfkeywords={Event Cameras, Stereo, Depth Estimation, 3D reconstruction}
}
\usepackage[absolute]{textpos}

\definecolor{firstcolor}{HTML}{BDE6CD}%
\definecolor{secondcolor}{HTML}{E2EEBC}%
\definecolor{thirdcolor}{HTML}{FFF8C5}%

\newcommand{\fst}[1]{\cellcolor{firstcolor}\bfseries#1}
\newcommand{\snd}[1]{\cellcolor{secondcolor}#1}
\newcommand{\trd}[1]{\cellcolor{thirdcolor}#1}

\usepackage{cuted}
\usepackage{capt-of}
\usepackage{caption}
\newcommand{\nh}[1]{\textcolor{black}{{#1}}}

\newcolumntype{Y}{>{\centering\arraybackslash}X}

\def\confName{CVPR}
\def\confYear{2026}

\title{Bidirectional Cross-Modal Prompting for Event-Frame Asymmetric Stereo\\ [-0.3cm]}

\author{
	Ninghui Xu$^{1,2,\dagger}$ \quad\quad Fabio Tosi$^{2}$ \quad\quad Lihui Wang$^{1,*}$ \quad\quad Jiawei Han$^{2,4}$ \quad\quad Luca Bartolomei$^{2}$ \\ \quad Zhiting Yao$^{3}$ \quad\quad Matteo Poggi$^{2}$ \quad\quad Stefano Mattoccia$^{2}$ \\ 
	[0.1cm] \small $^{1}$School of Instrument Science and Engineering, Southeast University, \\ [-0.1cm] \small State Key Lab of Comprehensive PNT Network and Equipment Technology, \\ [-0.1cm] \small Key Lab of Micro-Inertial Instrument and Advanced Navigation Technology, MOE \\ [-0.1cm]
	\small $^{2}$Department of Computer Science and Engineering, University of Bologna \\ [-0.1cm]
	\small $^{3}$College of Computer Science and Software Engineering, Hohai University \\ [-0.1cm]
	\small $^{4}$Beijing Institute of Technology \\ [-0.6cm]
}

\newcolumntype{C}[1]{>{\centering\arraybackslash}m{#1}}
\begin{document}
\definecolor{somegray}{gray}{0.5}
\newcommand{\darkgrayed}[1]{\textcolor{somegray}{#1}}
\begin{textblock}{11.5}(2.25, 0.8)  %
\begin{center}
\darkgrayed{This paper has been accepted for publication at the\\
IEEE Conference on Computer Vision and Pattern Recognition (CVPR), Denver, 2026.
\copyright IEEE}
\end{center}
\end{textblock}

\maketitle

\let\thefootnote\relax\footnotetext{
    \hspace{-0.5cm}$\dagger$ Work done while visiting the University of Bologna
    \newline *\ Corresponding author
}

\begin{strip}
	\centering
	\includegraphics[width=0.95\textwidth, page=1, trim=5 5 5 5, clip]{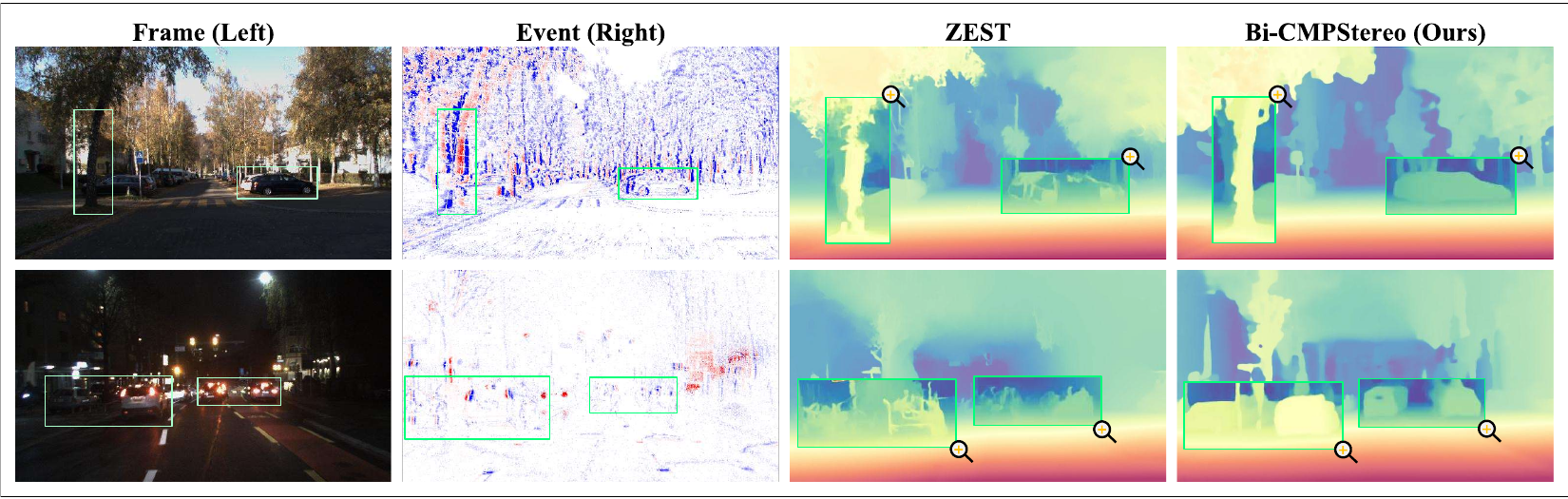}
	\vspace*{-0.7em}
	\captionof{figure}{\textbf{Qualitative comparison on event–frame asymmetric stereo.} Compared to the state-of-the-art ZEST~\cite{lou2024zero}, our method achieves higher accuracy and higher-quality structural details in both complex-texture (first row) and sparse low-light scenes (second row).}
	\label{fig:teaser}
\end{strip}

\begin{abstract}
Conventional frame-based cameras capture rich contextual information but suffer from limited temporal resolution and motion blur in dynamic scenes. Event cameras offer an alternative visual representation with higher dynamic range free from such limitations. The complementary characteristics of the two modalities make event-frame asymmetric stereo promising for reliable 3D perception under fast motion and challenging illumination. However, the modality gap often leads to marginalization of domain-specific cues essential for cross-modal stereo matching. In this paper, we introduce Bi-CMPStereo\footnote{\hspace{-0.5cm}Code is available at: \url{https://github.com/xnh97/Bi-CMPStereo}}, a novel bidirectional cross-modal prompting framework that fully exploits semantic and structural features from both domains for robust matching. Our approach learns finely aligned stereo representations within a target canonical space and integrates complementary representations by projecting each modality into both event and frame domains. Extensive experiments demonstrate that our approach significantly outperforms state-of-the-art methods in accuracy and generalization.
\end{abstract}
    
\section{Introduction}
\label{sec:intro}

Stereo matching aims to establish pixel-wise correspondences between stereo images, which is a fundamental technique in computer vision to compute dense disparity maps for depth estimation \cite{scharstein2002taxonomy}, with critical applications in robotics, autonomous driving, and AR/VR. Deep learning has driven remarkable progress for conventional RGB camera stereo in recent years \cite{tosi2024survey,laga2020survey,poggi2021synergies}, with iterative refinement-based methods demonstrating particular prominence \cite{lipson2021raft,li2022practical,wen2025stereo,xu2023iterative}.

As novel bio-inspired neuromorphic sensors, event cameras asynchronously detect per-pixel illumination changes (i.e., events) with microsecond temporal resolution \cite{lichtsteiner2008128}, embodying a paradigmatic shift from traditional visual modalities \cite{xu2023denoising}. By rapidly capturing the dynamic information, event-based sensors exhibit compelling advantages over frame-based vision: extremely low latency ($\leq$3 $\mu$s vs. 30 ms) and high dynamic range ($\geq$120 dB vs. 60 dB) \cite{xu2025mets, gallego2020event}. These properties have motivated their use in stereo matching for highly-dynamic scenarios with challenging illumination or high-speed motion \cite{nam2022stereo,zhang2022discrete,tulyakov2019learning,ghosh2025event}.

However, the inherent sparsity of events, which responds exclusively to temporal intensity variations, limits symmetric event stereo in extracting sufficient contextual information for dense disparity estimation \cite{ahmed2021deep,cho2022selection}. Moreover, the high cost of neuromorphic sensors further limits the deployment of dual-event systems in practical settings. Consequently, conventional frame-based cameras, offering rich contextual information and cost efficiency \cite{mostafavi2021event}, are complementary to event sensors, motivating growing interest in the stereo setups combining an event camera with a standard RGB sensor \cite{lou2024zero, wang2021stereo, ding2024video, zhang2022data, Chen_2024_WACV, cho2023non, lin2025learning}. Such stereo configurations with heterogeneous binocular inputs are defined as asymmetric systems, since they do not support a fully symmetric processing pipeline for the two views.

Despite the potential benefits of cross-modal asymmetric stereo, the significant modality gap between events and frames undermines the feature-space alignment assumption underlying stereo models to infer cross-view feature similarities. Several attempts have been made to mitigate the discrepancy through either 1) domain-level alignment that unifies events and frames into a common representation for a Siamese feature extractor \cite{lou2024zero,wang2021stereo,kim2022real} or 2) feature-level alignment employing separate extractors for each domain to derive shared embeddings \cite{ding2024video,cho2023non,lin2025learning}. However, focusing on cross-modal commonalities can marginalize discriminative domain-specific features that are salient in one domain but sparsely captured in the other. For instance, color cues readily available in images require complex networks to be extracted from events, making them prone to marginalization during alignment. This trade-off limits these models from achieving performance competitive with event symmetric counterparts using homogeneous binocular inputs. Therefore, the key challenge is learning expressive representations without information-lossy marginalization.

In this paper, we address this challenge by proposing Cross-Modal Prompting Stereo (CMPStereo) network, a novel network that preserves domain-specific semantic and structural cues crucial for robust stereo matching. Our key idea is to alternately designate the asymmetric stereo inputs as the target-domain and source-domain modalities, using the target domain as the canonical space for feature alignment. To this end, we introduce a Stereo Canonicalization Constraint (SCC) that regularizes the network to learn target-domain discriminative features from both modalities, promoting high-fidelity cross-modal alignment. A Cross-Domain Embedding Adapter (CDEA) further reinforces target-domain cues weakly encoded in source-domain representations. Through two complementary CMPStereo variants with events and images as target domains, we present Bidirectional Cross-Modal Prompting (Bi-CMPStereo) framework that simultaneously exploits bidirectional cost volumes across domains to achieve robust asymmetric stereo. To prevent shortcut learning in context features, we adopt Hierarchical Visual Transformation (HVT) \cite{yang2024learning} for enhanced robustness and generalization.

\cref{fig:teaser} shows the compelling performance of our proposals, and our contributions can be summarized as follows:

\begin{itemize}
    \item We propose CMPStereo with SCC and CDEA to preserve domain-specific features and achieve high-fidelity cross-modal alignment in a target canonical space. %
    
    \item We introduce Bi-CMPStereo, a bidirection framework integrating complementary information from event and image domains to enable robust disparity estimation.

    \item Our method significantly outperforms state-of-the-art approaches in both accuracy and generalization on DSEC and MVSEC benchmarks.
    
\end{itemize}

\section{Related Work}
\textbf{Frame-based Stereo.} Stereo matching has evolved from classical hand-crafted methods~\cite{scharstein2002taxonomy} to deep learning paradigms \cite{poggi2021synergies, tosi2024survey, laga2020survey, gong2025ov3r}. Early CNN-based networks~\cite{zbontar2015computing,luo2016efficient} replaced individual pipeline components, while end-to-end architectures diverged into 2D~\cite{mayer2016large, liang2018learning_iResNet,yin2019hierarchical,xu2020aanet} and 3D approaches~\cite{kendall2017end_GC-NET,chang2018pyramid,shen2021cfnet,zhang2019ga} processing cost volumes through 3D convolutions. RAFT-Stereo~\cite{lipson2021raft} pioneered iterative refinement with correlation volume-based updates, followed by subsequent works~\cite{xu2023iterative,li2022practical,wang2024selective,chen2024mocha,zhao2023high,xu2024igev++}. Transformer-based architectures~\cite{guo2022context_CEST,Li_2021_ICCV_STTR,xu2023unifying,min2025s2m2} leverage self-attention for long-range dependencies, while~\cite{guan2025bridgeDepth} propose data-driven MRF models. To improve generalization, various strategies have been explored, including domain-invariant feature learning~\cite{zhang2019domaininvariant,Zhang_2022_CVPR,Chang_2023_CVPR,Liu_2022_CVPR,Chuah_2022_CVPR}, geometric priors~\cite{aleotti2021neural,tosi2024neural}, and self-supervised learning with photometric losses \cite{godard2017unsupervised,Wang_2019_CVPR,Poggi_2024_CVPR} or pseudo-labels~\cite{tonioni2017unsupervised,poggi2021continual,Tosi_2023_CVPR}. Recently, foundation stereo models~\cite{wen2025stereo,jiang2025defom,cheng2025monster,Bartolomei2025StereoAnywhere,yao2025diving} combine pre-trained monocular depth with stereo matching, achieving superior generalization through large-scale diverse training.

\begin{figure*}[t]
	\centering
	\includegraphics[width=0.96\textwidth, page=2, trim=5 5 5 15, clip]{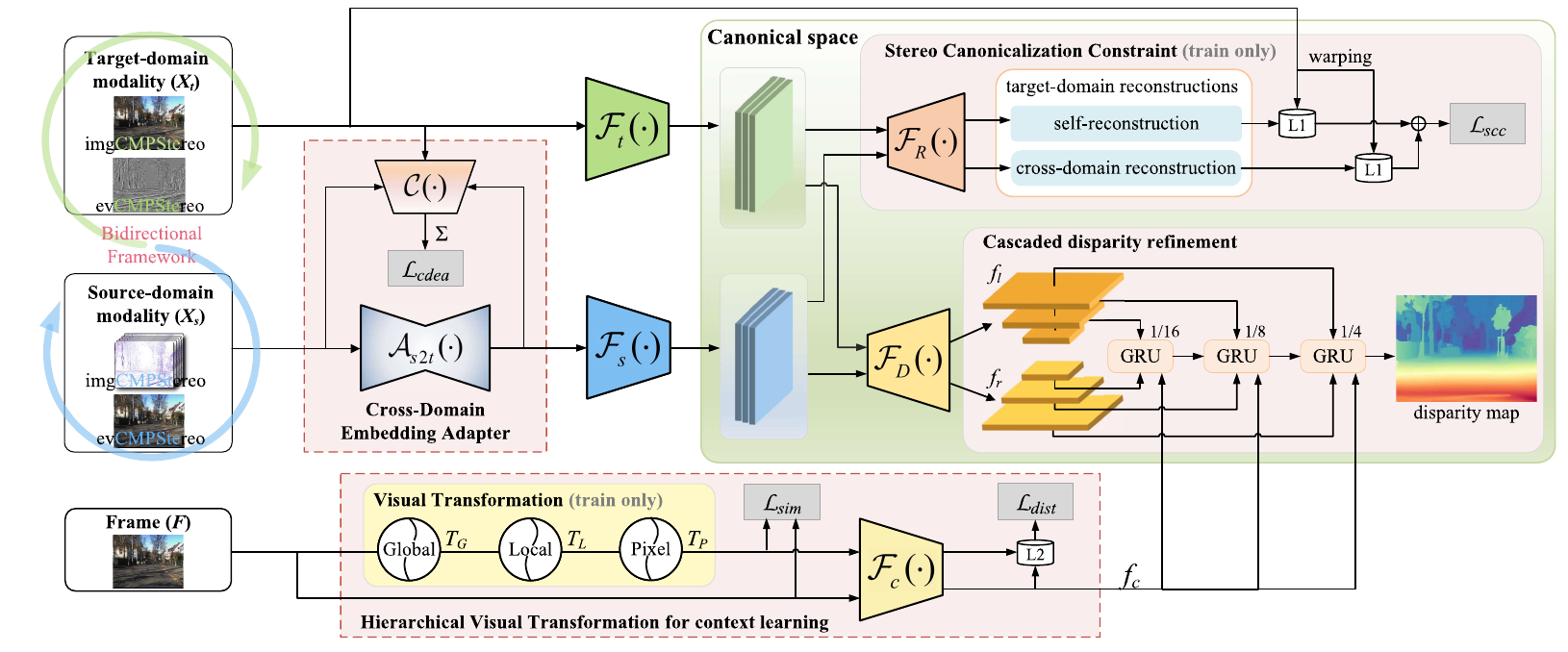}
    \vspace*{-1em}
	\caption{\textbf{Overview of the proposed CMPStereo network}. In our bidirectional framework, the asymmetric stereo inputs are alternately designated as the target-domain and source-domain modalities. The Cross-Domain Embedding Adapter (CDEA) operates on the source domain to achieve initial source-to-target adaptation. CMPStereo employs domain-specific encoders $\mathcal{F}_t(\cdot)$ and $\mathcal{F}_s(\cdot)$ combined with Stereo Canonicalization Constraint (SCC) to map the source and target domains onto a canonical space of the target domain. The multi-scale stereo features produced by a shared decoder $\mathcal{F}_D(\cdot)$ are fed to cascaded ConvGRUs for iterative disparity refinement. The context is extracted from the image frame $F$ with the Hierarchical Visual Transformation (HVT) \cite{yang2024learning} to avoid shortcut-learning.}
    \label{fig1}
    \vspace*{-4.5em}
\end{figure*}

\textbf{Event-based Stereo.} Symmetric event stereo leverages binocular event cameras to estimate disparity. Early studies~\cite{tulyakov2019learning,zhang2022discrete} modeled the asynchronous nature of events through temporal convolutions, with~\cite{zhang2022discrete} introducing continuous and discrete time operations to capture intrinsic dynamics. Recent works exploit the high temporal resolution of events via temporal aggregation: ~\cite{nam2022stereo} learns from past and future events to produce compact representations, while~\cite{cho2024temporal} introduces stereoscopic flow to continuously exploit information across timesteps, and~\cite{zhang2025ematch} unifies event-based flow and stereo in a shared correspondence space. However, event sparsity in static or low-texture regions hinders dense disparity estimation. To overcome this, ~\cite{ahmed2021deep,ghosh2024two} reconstruct image features from events to estimate dense disparity maps, while others transfer knowledge from image to event domains through unsupervised adaptation~\cite{cho2023learning} or cross-sensor distillation~\cite{zhang2025enhanced}. Complementary efforts integrate additional sensing modalities, such as LiDAR~\cite{bartolomei2024lidar} or active projection~\cite{Li_2025_CVPR}. Despite notable progress, symmetric event stereo remains constrained by inherent sparsity, motivating asymmetric setups that combine events with frames.

\textbf{Cross-Modal Event-Frame Stereo.} Cross-modal stereo leverages the complementary strengths of events and frames. Event-frame fusion stereo employs binocular pairs of both modalities, using sequential fusion ~\cite{mostafavi2021event}, differentiable event selection with cross-similarity~\cite{cho2022selection}, and cross-modality propagation~\cite{cho2022event} to complement sparse event features with dense frame information. However, requiring both modalities per view increases complexity and cost. Asymmetric stereo alleviates this by deploying an event camera in one view and a frame camera in the other, but the cross-view modality gap significantly intensifies matching challenges. Early handcrafted approaches~\cite{wang2021stereo, kim2022real} rely on edge-based matching, while~\cite{Chen_2024_WACV} decomposes the problem via temporal fusion and~\cite{zhang2022data} proposes a two-stage learning framework for event–frame association. \nh{More recently, Zhuang \etal~\cite{zhuang2025asymmetric} address the domain gap by gating frame features with temporal cues from frame sequences,} and~\cite{lou2024zero} achieves zero-shot generalization through visual prompting. However, existing strategies risk marginalizing discriminative domain-specific cues essential for stereo matching. Other recent works exploit event–frame disparity as an intermediate representation for tasks such as video interpolation~\cite{ding2024video} and motion deblurring~\cite{lin2025learning, cho2023non}.

\label{sec:related}

\section{Method Overview}
\label{sec:method}

We aim to estimate disparity from a calibrated event-frame asymmetric stereo setup. To address the inherent perceptual gap between the two sensing modalities, we introduce Bi-CMPStereo, a bidirectional framework that enforces semantic and structural consistency across domains. 
At its core, CMPStereo learns aligned stereo representations by mapping both target modality $X_t$ and source modality $X_s$ into a canonical space of the target domain for disparity estimation. To exploit complementary cues, we instantiate this architecture in two configurations that alternate the domain roles: evCMPStereo, which treats events (represented as event concentration $E$ \cite{nam2022stereo}) as $X_t$ and the frame $F$ as $X_s$, and imgCMPStereo, which reverses this assignment using $F$ as $X_t$ and events (encoded as voxel grid $V$ \cite{zhu2019unsupervised}) as $X_s$, following established practices in event-based vision. The key components of CMPStereo are detailed below.

\subsection{CMPStereo}
The overall architecture of CMPStereo is illustrated in Fig.~\ref{fig1}. The network employs domain-specific encoders to extract stereo features, which are aligned in the target canonical space through the Cross-Domain Embedding Adapter (CDEA) and Stereo Canonicalization Constraint (SCC). Context features are learned via Hierarchical Visual Transformation (HVT) to prevent shortcut learning, and disparity is iteratively refined through cascaded ConvGRUs.

\begin{figure*}[t]
	\centering
	\includegraphics[width=0.93\textwidth, page=3, trim=5 5 5 12, clip]{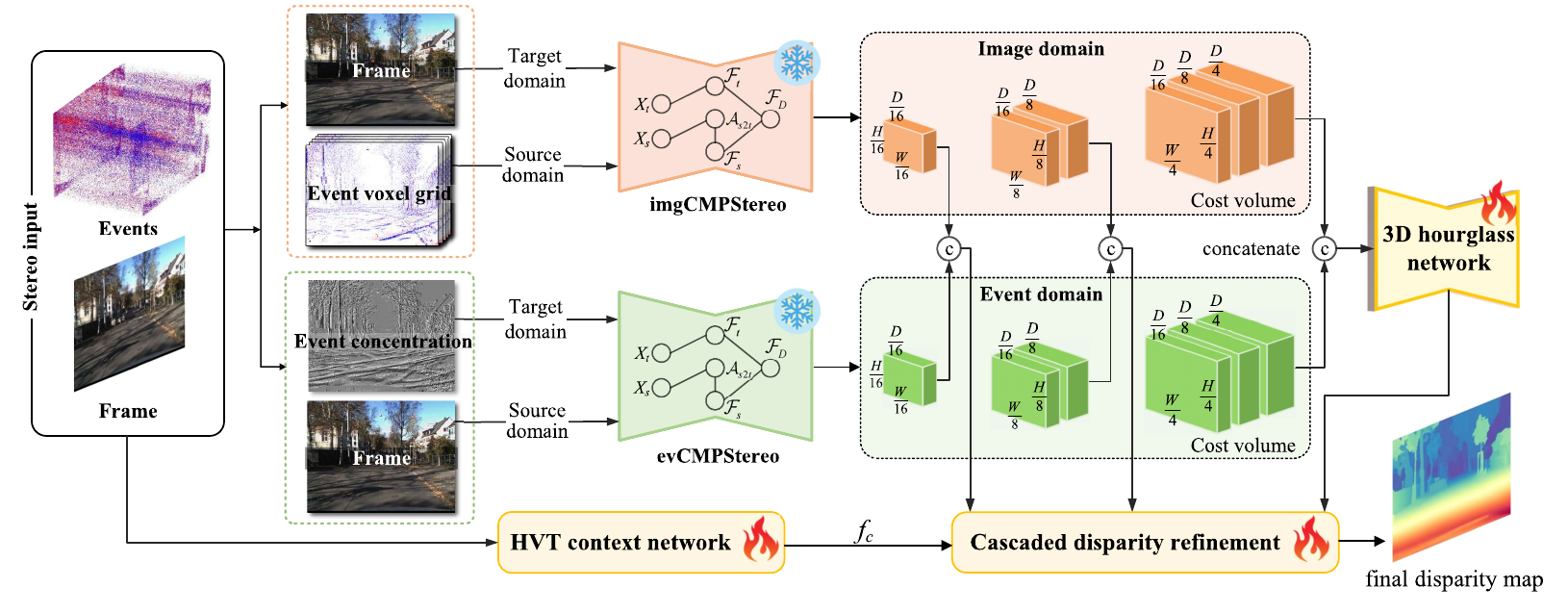}
    \vspace{-1.3em}
	\caption{\textbf{Overview of the Bi-CMPStereo framework.} Bi-CMPStereo integrates complementary representations from imgCMPStereo and evCMPStereo in the image and event domains for reliable disparity estimation. The two pre-trained CMPStereo networks are frozen as stereo feature extractors to construct multi-scale cost volumes, which are concatenated and fused via a 3D hourglass network at $1/4$ scale, followed by the same cascaded architecture for iterative disparity refinement.}
	\label{fig2}
    \vspace*{-4.3em}
\end{figure*}

\textbf{Cross-Domain Embedding Adapter.}
CMPStereo begins with the Cross-Domain Embedding Adapter (CDEA) for the source-domain modality $X_s$, which employs a U-shaped \cite{ronneberger2015u} adapter $\mathcal{A}_{s2t}(\cdot)$ to map $X_s$ into an embedding space aligned with the target domain. This process explicitly activates discriminative target-domain cues latent in the source representations, facilitating fine-grained feature alignment in the target domain.

To guide the adapter in modeling the target-domain embedding distribution and ensure the independence of two domain-specific CMPStereo variants, we employ a domain classifier $\mathcal{C}(\cdot)$ that distinguishes between event and frame embeddings. The classifier learns discriminative cues from both domains and supervises the source-to-target adaptation by minimizing the following loss:
\begin{equation}
	\begin{split}
		\mathcal{L}_{cdea} 
		&= \ell_{ce}(\mathcal{C}(E),1) + \ell_{ce}(\mathcal{C}(F),0) \\
		&\quad + \ell_{ce}(\mathcal{C}(\mathcal{A}_{s2t}(X_s)), \mathcal{Y}_t)
	\end{split}
\end{equation}
where $\ell_{ce}$ denotes the binary cross-entropy loss, $E$ and $F$ are event and frame representations, and $\mathcal{Y}_t$ is the target domain label. By using a single shared domain classifier, each adapter  learns to map its source domain specifically toward its designated target domain, ensuring that evCMPStereo and imgCMPStereo learn complementary representations in their respective canonical spaces.%

Notably, CDEA performs domain-level adaptation rather than explicit pixel-level translation, as the latter typically results in blurred representations \cite{rebecq2019high, mostafavi2021learning, messikommer2022bridging} that may diminish stereo-specific features. In addition, CDEA is designed based on a single domain classifier, which avoids adversarial learning and results in better training stability.

\textbf{Stereo Canonicalization Constraint.}
To extract cross-modal stereo representations, CMPStereo adopts two domain-specific encoders $\mathcal{F}_t(\cdot)$ and $\mathcal{F}_s(\cdot)$ to account for the modality discrepancy between the source and target domains, while employing a shared decoder $\mathcal{F}_D(\cdot)$ to accommodate the feature-space alignment assumption underlying stereo matching, producing multi-scale features $f_l^i,f_r^i \in \mathbb{R}^{C_i \times \frac{H}{i} \times \frac{W}{i}}, i \in \{4,8,16\}$ for cascaded disparity estimation. %
While stereo matching over the correlation volume implicitly drives the encoders to converge on a canonical latent space through joint optimization, the latent space is prone to collapsing into overly similar representations that marginalize discriminative cues. To mitigate this issue, we introduce the Stereo Canonicalization Constraint (SCC) in the bottleneck layer to achieve high-fidelity cross-domain alignment in the target canonical space.

Our goal is to train the domain-specific encoders to learn a canonical latent space that faithfully represents both modalities without marginalizing discriminative cues. The key insight is that an expressive intermediate representation should be able to faithfully reconstruct the original inputs in the target-domain space. %
We therefore enforce this reconstruction constraint during training using a shared lightweight decoder $\mathcal{F}_R(\cdot)$ %
that maps both source and target representations from the canonical space to their target-domain reconstructions:
\begin{equation}
	\begin{split}
		\mathcal{L}_{scc} 
		& = \| \mathcal{F}_R(\mathcal{F}_s(\mathcal{A}_{s2t}(X_s)) - X_s^{(t)}\|_1  \\
		&\quad + \| \mathcal{F}_R(\mathcal{F}_t(X_t) - X_t\|_1
	\end{split}
\end{equation}
where $X_s^{(t)}\!\!:=\!\!\mathcal{W}(X_t, d_{gt})$ is the source-domain modality expressed in the target-domain space, obtained by warping the target-domain counterpart \nh{using the ground-truth disparity $d_{gt}$. The deliberately lightweight design of $\mathcal{F}_R$ prevents it from hallucinating missing details, compelling the encoders to preserve fine-grained cues in the latent space.}

This constraint provides three-fold benefits: (1) the self-reconstruction of the target-domain modality preserves domain-specific discriminative features essential for stereo matching; (2) the cross-domain reconstruction of the source-domain modality enforces fine-grained source-to-target alignment, driving CMPStereo to operate within the target space; and (3) the shared reconstruction decoder effectively regularizes both domain-specific encoders toward a unified canonical latent space, promoting stereo-consistent feature generation. It is worth noting that SCC is employed only in the training stage to promote stereo representation learning while not involved during inference, thus imposing no additional runtime overhead at test time.

\textbf{Hierarchical Visual Transformation for context learning.}
Context features are essential for the ConvGRU updater, serving both to initialize its hidden state with a global scene prior and to inform each iteration \cite{teed2020raft}. We extract context features from the image frame, %
which provides rich semantic content, using a context network $\mathcal{F}_c(\cdot)$ with residual blocks and downsampling layers \cite{lipson2021raft}, producing multi-scale context features $f_c^i \in \mathbb{R}^{C \times \frac{H}{i} \times \frac{W}{i}}, i \in \{4,8,16\}$. However, in cross-modal stereo, the network may %
over-rely on the frame context to circumvent challenging cross-modal alignment, learning shortcut heuristics rather than exploiting stereo correspondenc, resulting in poor generalization.%

To alleviate this issue, we adopt the Hierarchical Visual Transformation (HVT) \cite{Chang_2023_CVPR, yang2024learning} to learn shortcut-invariant context features and facilitate the generation of robust stereo representations. HVT synthesizes a spectrum of augmented views by applying three hierarchical visual transformations $\{T_G(F), T_L(F), T_P(F)\}$ to the original frame $F$ from three levels: Global, Local, Pixel, enforcing features invariance across views to prevent superficial shortcuts. To ensure that the transformations induce substantial visual discrepancy, the visual similarity between original and transformed frames is minimized:
\begin{equation}
	\mathcal{L}_{sim} = \sum\limits_J {{\mathop{\rm Cos}\nolimits} ( {\phi ( {{T_J}( F )} ),\phi( F )} )}
\end{equation}
where $J \in \{G, L, P\}$ and $\rm Cos(\cdot,\cdot)$ denotes the cosine similarity function. $\phi(\cdot)$ extracts low-level visual features to measure pixel-level similarity.

Consistency between context features of transformed and original frames is enforced by minimizing:
\begin{equation}
	\mathcal{L}_{dist} = \sum\limits_J {{{\| {{\mathcal{F}_c}( {{T_J}( F)}) - {\mathcal{F}_c}( F )} \|}_2}}
\end{equation}
The combined objective is defined as:
\begin{equation}
	\mathcal{L}_{HVT} = \lambda_{hvt,1} \mathcal{L}_{sim} + \lambda_{hvt,2} \mathcal{L}_{dist}
\label{eq1}
\end{equation}
where $\lambda_{hvt,1}$ and $\lambda_{hvt,2}$ are loss weights. This objective guides the model to learn shortcut-invariant context features for robust stereo matching. Similar to SCC, HVT is applied solely as a training-time regularization on the context network without affecting inference efficiency.

\textbf{Cascaded Disparity Refinement.}
We estimate disparity through a cascaded coarse-to-fine architecture. At coarse resolutions, large receptive fields capture abstract scene structures with higher cross-modal coherence, yielding robust matching. To leverage this coarse-level robustness while exploiting fine-level details, the framework mitigates high-resolution ambiguity via coarse disparity priors and enables hierarchical cross-modal alignment: \nh{semantic consistency enforced at the coarse scale---where the modality gap is narrower---propagates through the shared decoder to yield fine-grained structural alignment at higher resolutions.}

Given multi-scale stereo features $f_l^i, f_r^i, i \!\in\!\{4,8,16\}$, we construct the cost volumes $C_{gwc}^i \in \mathbb{R}^{N_c \times \frac{D}{i} \times \frac{H}{i} \times \frac{W}{i}}$ at each scale using group-wise correlation \cite{guo2019group}:\begin{equation}
	C_{gwc}^i(g,d,x,y) \!=\! \frac{1}{{{N_c}/{N_g}}}\left\langle {f_{l,g}^i({x,y}),f_{r,g}^i({x\!-\!d,y})} \right\rangle
\end{equation}
where $N_c$ is the feature channels; $\left\langle\cdot,\cdot\right\rangle$ denotes the inner product; $g\in\{1,2,\dots\!,N_g\}$ is the group index among the total $N_g\!\!=\!\!8$ groups; $d\in\{1,2,\dots\!,\frac{D}{i}\}$ is the disparity index.

To further capture matching cues across scales, we construct multi-layer pyramids by pooling each cost volume: 3-, 2-, and 1-layer pyramids are built for the  1/4, 1/8, and 1/16 scales, respectively, all downsampled to the common resolution of 1/16, as shown in Fig.~\ref{fig2}. \nh{At each cascade level, a single-level ConvGRU iteratively updates the hidden state through lookups on the corresponding pyramid. The hidden state is initialized from context features and in each iteration a convolutional projection from the hidden state produces a residual disparity to update the current estimate.}

The disparity estimate from the previous scale is propagated as the initialization for the next cascade level via convex upsampling \cite{teed2020raft}, and the final prediction is similarly upsampled to the full resolution.

\newlength{\wMethod} \setlength{\wMethod}{2.48cm}
\newlength{\wNum}    \setlength{\wNum}{0.92cm}
\newlength{\wPE}    \setlength{\wPE}{1.03cm}
\newlength{\wPEl}    \setlength{\wPEl}{1.06cm}
\newlength{\wPEll}    \setlength{\wPEll}{1.1cm}
\newcommand{\gsep}{0.0cm}
\setlength{\tabcolsep}{2pt}
\begin{table*}[t]
	\centering
    \resizebox{\textwidth}{!} {
	\begin{tabular}{@{}%
			C{\wMethod}%
			@{\hspace{\gsep}}*{4}{C{\wNum}}%
			@{\hspace{\gsep}}*{4}{C{\wPEll}}%
            @{\hspace{\gsep}}*{1}{C{\wPEl}}%
            @{\hspace{\gsep}}*{3}{C{\wPE}}%
			@{\hspace{\gsep}}*{4}{C{\wNum}}@{}}
		\toprule
		\multirow{2}{*}{Method} &
		\multicolumn{4}{c}{MAE $\downarrow$} & \multicolumn{4}{c}{1PE $\downarrow$} & \multicolumn{4}{c}{2PE $\downarrow$} & \multicolumn{4}{c}{RMSE $\downarrow$} \\
		\cmidrule(lr){2-5}\cmidrule(lr){6-9}\cmidrule(lr){10-13}\cmidrule(lr){14-17}
		& Zu & In & Th & All & Zu & In & Th & All & Zu & In & Th & All & Zu & In & Th & All \\[-2pt]
  
		\midrule
		\multicolumn{17}{c}{Event-frame asymmetric stereo} \\
		\midrule

		ZEST\cite{lou2024zero} & 2.483 & 1.282 & 1.339 & 2.168 & 48.325 & 39.415 & 38.127 & 46.920 & 20.195 & 13.847 & 13.303 & 19.288 & 3.779 & 2.319 & 2.608 & 3.394 \\[1pt]

        ZEST$^{\dag}$\cite{lou2024zero} & 0.785 & 0.580 & 0.797 & 0.763 & 21.003 & 12.913 & 19.922 & 20.382 & 4.845  & 3.303  & 4.615  & 4.646  & 1.474 & 1.211 & 1.713 & 1.438 \\[1pt]
        
        SEVFI \cite{ding2024video} & 0.729 & 0.623 & 0.875 & 0.711 & 17.374 & 14.558 & 24.225 & 16.932 & 4.497  & 3.408  & 5.970  & 4.307  & 1.552 & 1.309 & 1.844 & 1.509 \\[1pt]

        \hdashline
        
        \textbf{evCMPStereo}   &  \trd{0.594}  &  \trd{0.496} &  \trd{0.761}  & \trd{0.577} &  \trd{12.752} & \trd{10.031} & 18.508 & 12.309 & \trd{3.097} & \trd{2.060} & \trd{4.116} & \trd{2.909} & \trd{1.352} & \snd{1.070} & \snd{1.697} & \trd{1.310} \\[2pt]
  
		\textbf{imgCMPStereo}  & \snd{0.582} & \snd{0.482} & \snd{0.747} & \snd{0.565} & \snd{11.989} & \snd{8.706} & \snd{17.530} & \snd{11.432} & \snd{3.000} & \snd{1.852} & \snd{3.977}  & \snd{2.790} & \snd{1.324} & \trd{1.132} & \trd{1.703} &  \snd{1.292}\\[2pt]
  
		\textbf{Bi-CMPStereo}  &\fst{0.547} & \fst{0.450} & \fst{0.715} &  \fst{0.532} &  \fst{11.112} & \fst{7.992} & \fst{16.444}  & \fst{10.613} &  \fst{2.606} & \fst{1.507} & \fst{3.587} & \fst{2.415} & \fst{1.244} & \fst{1.026} & \fst{1.651} & \fst{1.210}\\[-2pt]
        
        \midrule
		\multicolumn{17}{c}{Event-based symmetric stereo} \\
		\midrule
  
		SE-CFF \cite{nam2022stereo} &  0.627  &  0.555  &  0.771 &  0.612  & 13.123  & 10.397 & 17.745  &  12.477  & 3.506 &  2.536  & 4.408  & 3.288  & 1.462 & 1.325  & 1.862  & 1.445 \\
  
		DTC \cite{zhang2022discrete} &  0.631  &  0.563  &  0.768 &  0.621  & 12.809  & 10.183 & \trd{17.569}  &  \trd{12.069}  & 3.153 &  2.481  & 4.129  & 3.018  & 1.501 & 1.357  & 1.736  & 1.508 \\[-2pt]
  
		\bottomrule
	   \end{tabular}}\vspace{-0.2cm}
     \caption{\textbf{Comparison of our method with state-of-the-art stereo methods on the DSEC dataset \cite{gehrig2021dsec}.} Zu, In, and Th denote the Zurich City, Interlaken, and Thun sequences in the dataset, respectively. }
     \vspace{-4.6em}
	\label{tab1}
\end{table*}

\subsection{Bi-CMPStereo}
Within Bi-CMPStereo framework, evCMPStereo and imgCMPStereo are first trained independently end-to-end. We use the Smooth L1 loss with exponentially increasing weights
to compute the disparity loss $\mathcal{L}_{d}$ on all updated disparities $\{d_i\}_{i=1}^N$ at all scales:
\begin{equation}
	\mathcal{L}_{d} = \sum\limits_{i = 1}^N {{\gamma ^{N - i}}{{\left\| {{d_i} - {d_{gt}}} \right\|}_1}}
 \label{eq4}
\end{equation}

The overall training objective for CMPStereo is defined as a weighted combination of all losses:
\begin{equation}
	\mathcal{L}_{pre} = \mathcal{L}_{d} + \lambda_{1}\mathcal{L}_{cdea} + \lambda_{2}\mathcal{L}_{scc} + \lambda_{3}\mathcal{L}_{HVT}
 \label{eq8}
\end{equation}
where $\lambda_{1}$, $\lambda_{2}$, and $\lambda_{3}$ are loss weights.

After training, both networks learn well-aligned stereo representations within their respective domains. To leverage complementary features from both modalities, Bi-CMPStereo integrates these representations in a unified network, as shown in Fig.~\ref{fig2}. Specifically, for a given scene, each modality is processed by the pre-trained frozen evCMPStereo and imgCMPStereo networks, producing intra- and cross-modal representations to construct two complementary cost volumes at multiple scales. Subsequently, at $1/16$ and $1/8$ scales, the two cost volumes are directly concatenated to efficiently combine dual-modal information. At $1/4$ scale, they are fused through a 3D hourglass network \cite{xu2022attention} to aggregate complementary matching cues. The resulting multi-scale cost volumes, together with context features from HVT, are used for cascaded disparity refinement, as in CMPStereo. %
In this stage, only the 3D hourglass network, the ConvGRU updater, and the HVT context network are optimized with:
\begin{equation}
	\mathcal{L}_{final} = \mathcal{L}_{d} + \lambda_{3}\mathcal{L}_{HVT}
\label{eq3}
\end{equation}

\section{Experiments}
\label{sec:experiments}

\begin{figure*}[t]
	\centering
	\includegraphics[width=1\textwidth, page=4, trim=5 5 5 5, clip]{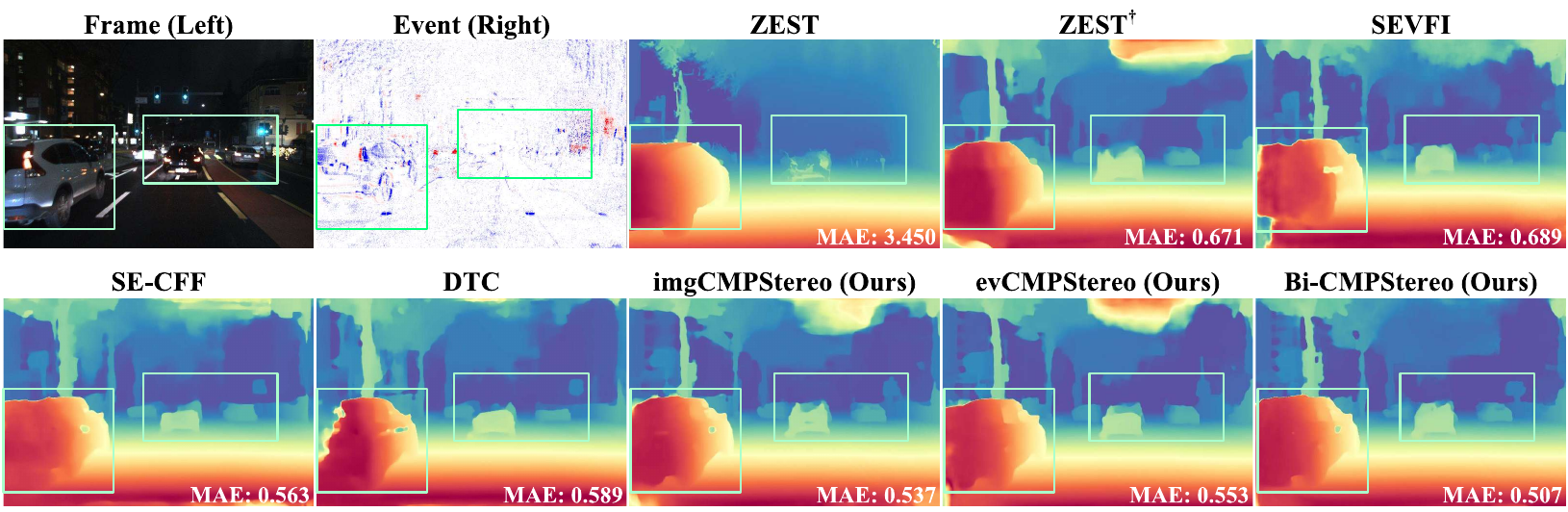}
	\vspace*{-2em}
	\caption{\textbf{Qualitative comparison with state-of-the-art methods on a low-light scene from the DSEC dataset \cite{gehrig2021dsec}.} The mean absolute error (MAE) is shown at the bottom right of each estimated disparity map. As can be seen in the highlighted regions, our method more accurately reconstructs the fine structural details of the cars.}
	\label{fig4}
	\vspace*{-4.5em}
\end{figure*}

\subsection{Implementation and Experimental Settings}
In this section, we describe our experimental setup including datasets, implementation details, evaluation metrics, and baseline methods.

\textbf{Datasets.}
We evaluate our method on \nh{DSEC \cite{gehrig2021dsec}, a high-quality event stereo dataset captured in outdoor driving scenarios}. It employs two Prophesee Gen 3.1 event cameras (640$\times$480) to provide stereo event streams, while synchronized intensity frames are captured by a separate pair of \nh{RGB cameras. DSEC contains 41 driving scenarios with 26K samples, covering diverse lighting and environmental conditions.} Following \cite{bartolomei2025depth}, we split the dataset into 31 training sequences with 19K samples and 10 testing sequences with 7K samples. For asymmetric stereo, we use the left intensity frame and the right event stream, where the intensity frames are warped to the left event camera using the provided calibration parameters. \nh{We also evaluate the generalization of our method on MVSEC \cite{zhu2018multivehicle} \nh{and M3ED \cite{chaney2023m3ed}} after training on DSEC \cite{gehrig2021dsec}. MVSEC utilizes two DAVIS346B (346×260) to provide stereo events and frames, and M3ED delivers synchronized data via Prophesee EVK4 HD event cameras (1280$\times$720). Following \cite{lou2024zero}, we evaluate on three MVSEC subsets: indoor\_flying1, 2, 3 (with frames 500-1500, 500-2000, and 500-2500, denoted as S1, S2, and S3), and one M3ED subset: car\_urban\_day\_horse (300-700).}

\begin{table}[t]
	\small
    \centering
	\begin{tabular}{
			p{2.8cm}  @{\hspace{0.03cm}}
			c c c c
		}
		\hline 
		Method  & MAE$\downarrow$ & 1PE$\downarrow$ & 2PE$\downarrow$ & RMSE$\downarrow$ \\ \hline
		SE-CFF \cite{nam2022stereo} & \trd{0.519} & 9.583 & 2.620 & \trd{1.231} \\[-0.5ex]
		DTC \cite{zhang2022discrete} & 0.527 & \trd{9.277} & \trd{2.416} & 1.291 \\[-0.5ex]
		Zhuang \etal \cite{zhuang2025asymmetric}& \snd{0.488} & \snd{8.881} & \snd{2.335} & \snd{1.168} \\[-0.5ex]
		\textbf{Bi-CMPStereo} & \fst{0.475} & \fst{8.557} & \fst{2.253} & \fst{1.101} \\ [-1.5pt] \hline
	\end{tabular}
	\vspace{-0.2cm}
	\caption{\textbf{Comparison on the DSEC online benchmark.}}
	\label{tab7}
	\vspace{-5.5em}
\end{table}

\textbf{Implementation Details.}
We implement our networks in PyTorch \cite{paszke2019pytorch} and use two NVIDIA RTX 3090 GPUs (each with 24 GB memory) for training and inference. We use the Adam \cite{kingma2014adam} optimizer with an initial learning rate of $5e^{-4}$, which decays following a cosine annealing schedule. We first pre-train evCMPStereo and imgCMPStereo independently for 100 epochs each with a batch size of 8, and then train the final Bi-CMPStereo network for another 100 epochs with the same batch size. The loss weights $\lambda_{hvt,1}$ and $\lambda_{hvt,2}$ in Eq~\ref{eq1} are set as 1 and 0.5 respectively. $\lambda_{1}$, $\lambda_{2}$ and $\lambda_{3}$ in Eq~\ref{eq8} and Eq~\ref{eq3} are set as 0.5, 2 and 1. $\gamma$ in Eq~\ref{eq4} is set to 0.9. %
On \nh{our DSEC training split} with 18948 samples, both evCMPStereo and imgCMPStereo take about 35 minutes per epoch, while the Bi-CMPStereo requires around 46 minutes. During inference, Bi-CMPStereo runs at about 0.1s per frame with the 640$\times$480 resolution.

\textbf{Metrics.}
We use the standard metrics for evaluation, including mean absolute error (MAE) and root mean square error (RMSE) of the disparity, as well as the n-pixel error (nPE) that is the percentage of pixels whose disparity error exceeds n pixels. We report 1PE and 2PE for DSEC, and 2PE and 3PE for \nh{MVSEC and M3ED} in the generalization evaluation. The {\setlength{\fboxsep}{0pt}\colorbox{firstcolor}{\strut first}}, {\setlength{\fboxsep}{0pt}\colorbox{secondcolor}{\strut second}} and {\setlength{\fboxsep}{0pt}\colorbox{thirdcolor}{\strut third}}-best are highlighted with different colors in tables.

\textbf{Baseline Methods.}
We compare our method with the state-of-the-art event-frame asymmetric stereo method ZEST \cite{lou2024zero}. ZEST aligns the representations of frames and events, enabling the use of foundation monocular \cite{yang2024depth} and stereo \cite{li2022practical} models pretrained on diverse image datasets for monocular cue-guided cross-modal stereo. To make ZEST more competitive on DSEC, we retrain it under the same training configuration as our method, denoted as ZEST$^{\dag}$. We also compare with SEVFI-Net \cite{ding2024video}, a deformable convolution-based network to estimate cross-modal stereo disparities. For a comprehensive assessment, we include two state-of-the-art event-based symmetric stereo models: SE-CFF \cite{nam2022stereo} and DTC-SPADE \cite{zhang2022discrete} as baseline methods.

\newlength{\wMethodo} \setlength{\wMethodo}{2.39cm}
\newlength{\wPEo}    \setlength{\wPEo}{1.1cm}
\setlength{\tabcolsep}{2pt}
\begin{table*}[t]
 \centering
 \resizebox{\textwidth}{!} {
	\begin{tabular}{@{}%
			C{\wMethodo}%
            @{\hspace{\gsep}}*{3}{C{\wNum}}%
			@{\hspace{\gsep}}*{1}{C{\wPEo}}%
			@{\hspace{\gsep}}*{4}{C{\wPEo}}%
			@{\hspace{\gsep}}*{4}{C{\wPEo}}%
			@{\hspace{\gsep}}*{4}{C{\wPEo}}@{}}
		\toprule
		\multirow{2}{*}{Method} &
		\multicolumn{4}{c}{MAE $\downarrow$} & \multicolumn{4}{c}{2PE $\downarrow$} & \multicolumn{4}{c}{3PE $\downarrow$} & \multicolumn{4}{c}{RMSE $\downarrow$} \\
		\cmidrule(lr){2-5}\cmidrule(lr){6-9}\cmidrule(lr){10-13}\cmidrule(lr){14-17}
		& S1 & S2 & S3 & All & S1 & S2 & S3 & All & S1 & S2 & S3 & All & S1 & S2 & S3 & All \\[-2pt]
		\midrule
		\multicolumn{17}{c}{Event-frame asymmetric stereo} \\
		\midrule
		ZEST\cite{lou2024zero}  & 3.340 & 7.190 & 5.130 & 5.220 & \snd{33.220} & 61.380 & \snd{43.970} & \snd{46.190} & \snd{19.060} & 41.720 & \snd{28.660} & \trd{29.820} & \snd{3.830} & 7.830 & \trd{5.600} & \trd{5.763} \\[1pt]
		
		ZEST$^{\dag}$\cite{lou2024zero} & 4.923 & 6.022 & 4.267 & 4.998 & \trd{44.676} & 60.123 & \trd{45.688} & 50.275 & 32.255 & 46.015 & 31.699 & 36.594 & 6.626 & 8.023 & 5.799 & 6.724 \\[1pt]
		
		SEVFI \cite{ding2024video}  & 5.537 & 4.728 & 5.779 & 5.375 & 63.568 & 60.108 & 68.703 & 64.697 & 48.482 & 45.459 & 53.147 & 49.548 & 10.143 & 8.352 & 10.440 & 9.678 \\[1pt]
		
		\textbf{Bi-CMPStereo}  & \fst{1.743} & \fst{2.181} & \fst{1.674} & \fst{1.858} & \fst{29.247} & \fst{37.041} & \fst{29.869} & \fst{32.121} & \fst{14.924} & \fst{20.889} & \fst{15.336} & \fst{17.095} & \fst{2.708} & \fst{3.419} & \fst{2.466} & \fst{2.837} \\[-2pt]
		\midrule
		
		\multicolumn{17}{c}{Event-based symmetric stereo} \\
		\midrule
		
		SE-CFF \cite{nam2022stereo} & \trd{2.428} & \trd{2.867}	& \trd{2.518}	& \trd{2.614}	& 52.993 & \trd{53.451} & 52.837 & 53.076	& 30.102 & \trd{32.864} & 32.238 & 31.972 & 5.729	& \trd{6.241}	& 5.807	& 5.934
		 \\[1pt]
		 
		DTC \cite{zhang2022discrete} & \snd{2.175} & \snd{2.502} & \snd{2.204} & \snd{2.297} & 48.767 & \snd{50.646} & 50.237 & \trd{50.047}	& \trd{27.382} & \snd{29.928} & \trd{29.530} & \snd{29.185} & \trd{5.313} & \snd{5.571} & \snd{5.210} & \snd{5.353}
		 \\[-2pt]
		
		\bottomrule
	\end{tabular} }\vspace{-0.3cm}
 	\caption{\textbf{Cross-dataset generalization results on the MVSEC~\cite{zhu2018multivehicle} dataset.} S1, S2 and S3 refers to single MVSEC subsets.}
	\label{tab2}
 \vspace{-4em}
\end{table*}

\subsection{Results on DSEC Dataset}
We compare our imgCMPStereo, evCMPStereo, and Bi-CMPStereo with baseline methods \nh{on DSEC \cite{gehrig2021dsec}}. Table~\ref{tab1} shows that our method consistently outperforms all baselines across all metrics. Notably, even our single-domain variants achieve superior performance, demonstrating the effectiveness of CMPStereo for cross-modal stereo. By integrating the complementary cues from both branches, Bi-CMPStereo achieves substantial further gains. It can be seen that existing asymmetric stereo methods remain inferior to event-based symmetric approaches, primarily because their alignment schemes suppress modality-specific cues. In contrast, our approach achieves high-fidelity alignment by effectively preserving discriminative features from each modality, surpassing even state-of-the-art event-based symmetric methods. Visual comparisons in Fig.~\ref{fig4} highlight the superior quality of our approach, producing disparity maps with sharper edges and richer scene content.

\nh{Furthermore, we evaluate our method on the official DSEC online benchmark, where ground-truth disparities are withheld. We additionally compare with Zhuang \etal \cite{zhuang2025asymmetric}, a recent asymmetric approach utilizing temporal feature gating from frame sequences. Since their source code is unavailable, we directly compare with their reported metrics. Results in Table~\ref{tab7} show that we outperform both symmetric baselines (SE-CFF~\cite{nam2022stereo}, DTC~\cite{zhang2022discrete}) and the multi-frame-based approach \cite{zhuang2025asymmetric}, further demonstrating the remarkable capability and robustness of our method.} 

\subsection{Cross-Dataset Generalization on MVSEC}
\begin{figure}[t]
	\centering
	\includegraphics[width=0.49\textwidth, page=5, trim=16 5 5 5, clip]{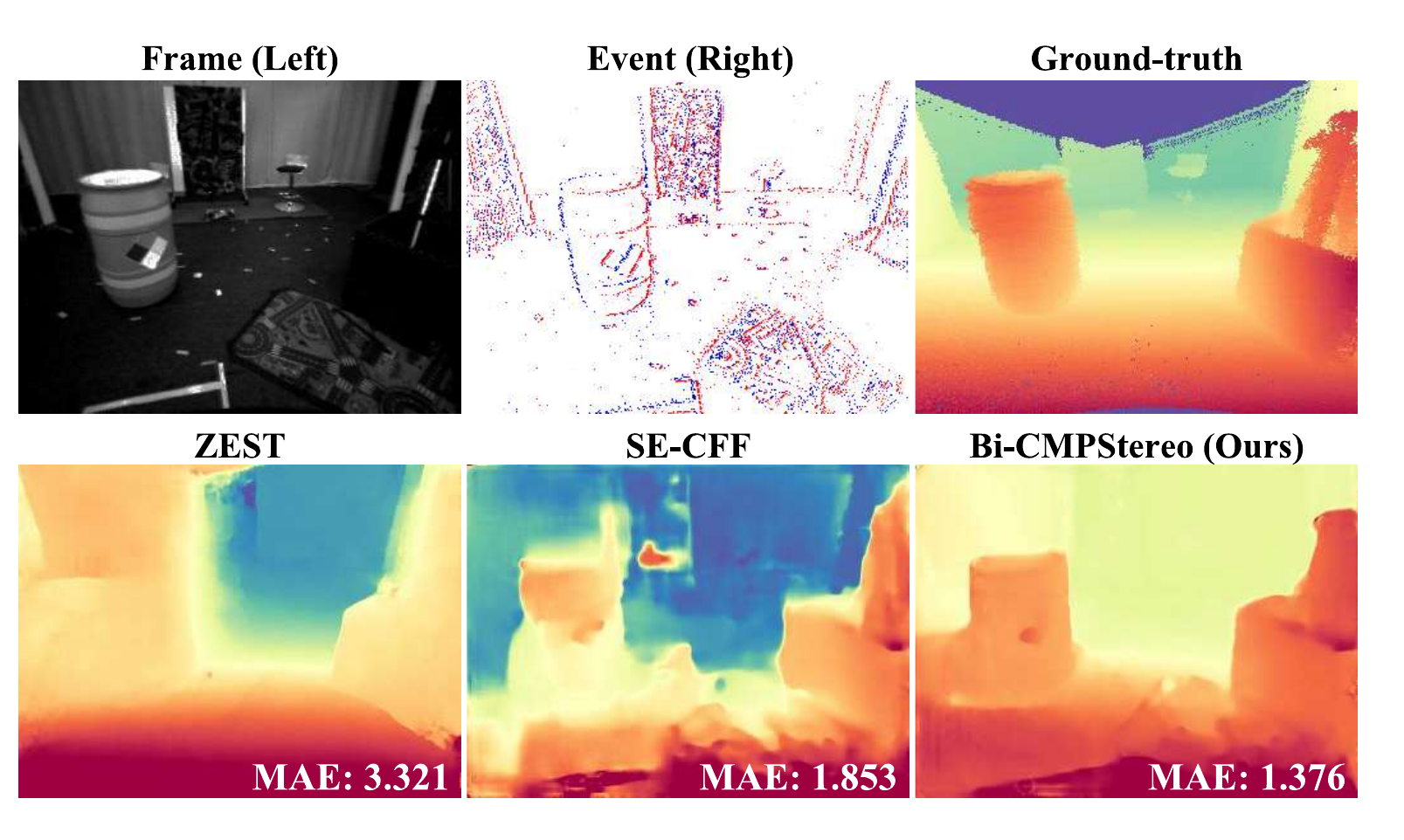}
	\vspace*{-2.5em}
	\caption{\textbf{Generalization Result}. Qualitative comparison of cross-dataset generalization on the MVSEC~\cite{zhu2018multivehicle} dataset.}
	\label{fig5}
	\vspace*{-4.5em}
\end{figure}

\nh{We further evaluate the generalization of our method by directly applying the DSEC-trained model to MVSEC~\cite{zhu2018multivehicle} and M3ED~\cite{chaney2023m3ed}}. \nh{We compare it against ZEST~\cite{lou2024zero}, which serves} as a representative baseline, as it is designed for zero-shot generalization via visual prompting from the image domain. \nh{As shown in Table~\ref{tab2} and Table~\ref{tab6}, our Bi-CMPStereo exhibits superior generalizability, consistently outperforming any competing methods. Leveraging models pretrained on large-scale image data, ZEST~\cite{lou2024zero} achieves competitive generalization, outperforming its DSEC-trained variant ZEST$^{\dag}$ in most cases. However, even with its external image-domain knowledge, ZEST still lags behind event-based symmetric baselines.} In contrast, our Bi-CMPStereo, trained solely on DSEC, achieves superior performance across all metrics, demonstrating the robust generalization to unseen scenes. Qualitative results in Fig.~\ref{fig5} further confirm the effectiveness of our method across datasets.

\begin{table}[t]
\small
\centering
\begin{tabular}{
    @{}
    p{1.8cm}  @{\hspace{0.1cm}}
    p{2cm}  @{\hspace{0.1cm}}
    c@{\hspace{0.3cm}} c c@{\hspace{0.2cm}} c@{}
}
\hline
 & Method & MAE$\downarrow$ & 2PE$\downarrow$ & 3PE$\downarrow$ & RMSE$\downarrow$ \\ \hline
\multirow{3}{=}{\linespread{0.5}\selectfont Event-frame Asymmetric Stereo}
& ZEST\cite{lou2024zero} & 2.060 & 29.020 & 19.040 & 3.390 \\[-0.5ex]
& ZEST$^{\dag}$\cite{lou2024zero} & 2.653 & 36.011 & 23.020 & 5.069 \\[-0.5ex]
& SEVFI\cite{ding2024video} & 3.634	& 52.107 & 36.615 & 5.282  \\[-0.5ex]
& \textbf{Bi-CMPStereo} & \fst{1.557} & \fst{17.044} & \fst{11.420} & \fst{2.877} \\ \hline 
\multirow{2}{=}{Event Stereo}
& SE-CFF\cite{nam2022stereo}& \snd{1.615} & \snd{17.364} & \snd{12.182} & \snd{2.976} \\ [-0.5ex]
& DTC\cite{zhang2022discrete} & \trd{1.902} & \trd{22.375} & \trd{15.661} & \trd{3.068} \\ [-1pt] \hline
\end{tabular}
\vspace{-0.3cm}
\caption{\textbf{Generalization results on the M3ED~\cite{chaney2023m3ed} dataset.}}
\label{tab6}
\vspace{-4em}
\end{table}

\subsection{Ablation Study}

\begin{table}[t]
\centering
\renewcommand{\arraystretch}{1} 
\begin{tabularx}{\linewidth}{
  p{3.16cm}YYYY
}
\hline
Method   & MAE$\downarrow$ & 1PE$\downarrow$ & 2PE$\downarrow$ & RMSE$\downarrow$  \\ \hline
w/o CDEA \& SCC   & 0.594 & 12.232 & 3.040 & 1.351 \\
w/o CDEA          & 0.583 & 12.118 & 2.968 & 1.325 \\
w/o SCC           & 0.589 & 12.054 & 2.945 & 1.354 \\
w/o cascades      & 0.588 & 11.927 & 2.994 & 1.347 \\
full imgCMPStereo & \fst 0.565 & \fst 11.432 & \fst 2.790 & \fst 1.292 \\ \hline
\end{tabularx}
\vspace{-0.3cm}
\caption{\textbf{Ablation study of imgCMPStereo.} Results on the  DSEC~\cite{gehrig2021dsec} dataset.}
\label{tab3}
\vspace{-4.5em}
\end{table}

In this section, we perform a series of ablation studies to validate the effectiveness of each component in our method. %

To assess the impact of each module on CMPStereo network, we evaluate imgCMPStereo with different settings on the DSEC dataset~\cite{gehrig2021dsec} and report the results in Table~\ref{tab3}. The w/o cascades variant performs disparity optimization only at the finest 1/4 scale with 10 iterations to maintain the same total iteration count. The drop in its accuracy demonstrates the importance of the cascaded architecture. Removing the CDEA module weakens modality alignment by failing to sufficiently activate potential target-domain cues in the source-domain modality, leading to a clear performance drop and confirming its effectiveness. The effectiveness of the SCC module is evident from the substantial accuracy gain it brings to CMPStereo, highlighting its pivotal role in promoting high-fidelity cross-modal alignment.

\begin{figure}[t]
	\centering
	\includegraphics[width=0.49\textwidth, page=6, trim=16 5 5 5, clip]{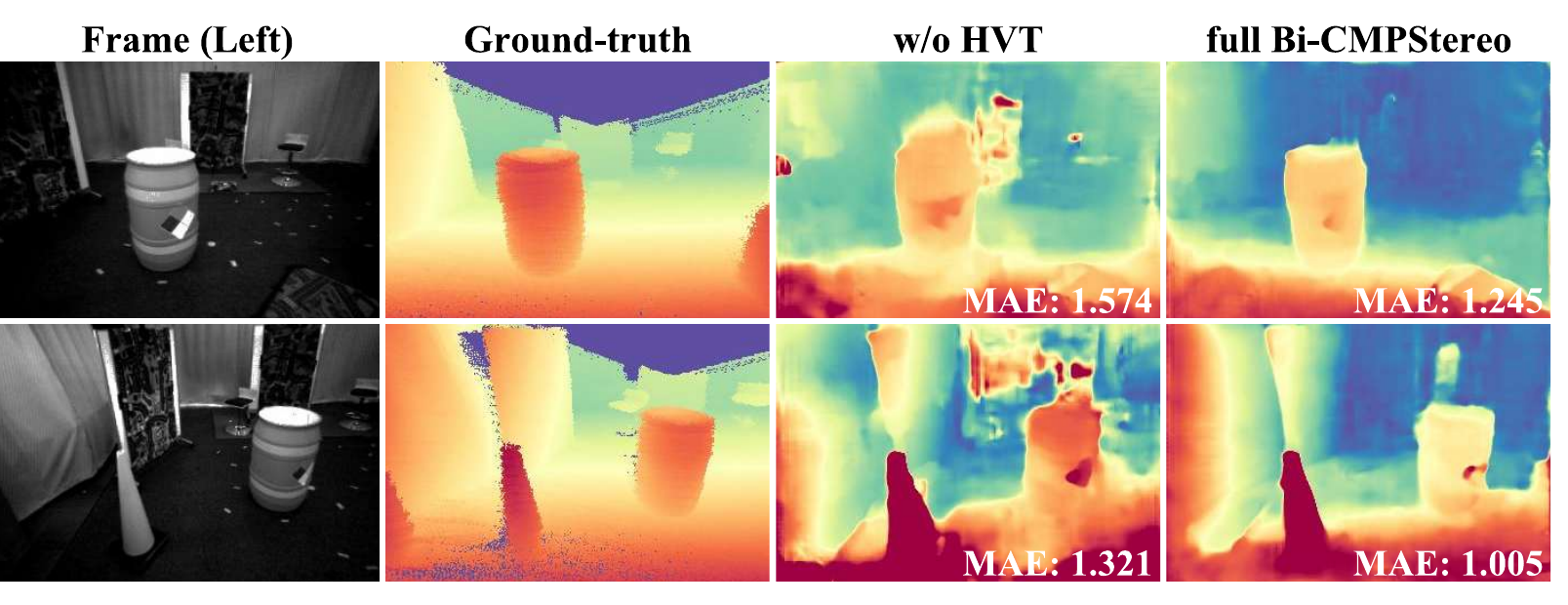}
	\vspace*{-2em}
	\caption{\textbf{Visual comparison between Bi-CMPStereo with and without HVT.} Generalization results on MVSEC~\cite{zhu2018multivehicle}.}
	\label{fig6}
	\vspace*{-4.5em}
\end{figure}

We validate the effectiveness of CDEA and SCC for bidirectional estimation in Table~\ref{tab4}, \nh{showing that they significantly enhance Bi-CMPStereo's performance.} Beyond facilitating modality alignment, CDEA and SCC further promote the two single-domain CMPStereo branches toward their respective target domains, thereby reinforcing their mutual complementarity and leading to more precise stereo predictions. HVT for context learning to enhance generalization is validated on Bi-CMPStereo through DSEC-to-MVSEC~\cite{zhu2018multivehicle} cross-dataset evaluation. As presented in Table~\ref{tab5}, HVT-based context learning achieves notable improvements in generalization, and the visual comparison in Fig.~\ref{fig6} substantiates its enhanced robustness across datasets.

\begin{table}[t]
\centering
\renewcommand{\arraystretch}{1} 
\begin{tabularx}{\linewidth}{
  p{3.16cm}YYYY
}
\hline
Method   & MAE$\downarrow$ & 1PE$\downarrow$ & 2PE$\downarrow$ & RMSE$\downarrow$  \\ \hline
w/o CDEA          & 0.546 & 11.128 & 2.610 & 1.239 \\
w/o SCC           & 0.551 & 11.290 & 2.614 & 1.257 \\
full Bi-CMPStereo & \fst 0.532 & \fst 10.613 & \fst 2.415 & \fst 1.210 \\ \hline
\end{tabularx}
\vspace{-0.3cm}
\caption{\textbf{Ablation study of Bi-CMPStereo}. Results on the DSEC~\cite{gehrig2021dsec} dataset.}
\label{tab4}
\vspace{-3.5em}
\end{table}

\begin{table}[t]
\centering
\renewcommand{\arraystretch}{1} 
\begin{tabularx}{\linewidth}{
  p{3.16cm}YYYY
}
\hline
Method   & MAE$\downarrow$ & 2PE$\downarrow$ & 3PE$\downarrow$ & RMSE$\downarrow$  \\ \hline
w/o HVT          & 2.093 & 36.005 & 22.478 & 3.060 \\
full Bi-CMPStereo & \fst 1.858 & \fst 32.121 & \fst 17.095 & \fst 2.837 \\ \hline
\end{tabularx}
\vspace{-0.3cm}
\caption{\textbf{Impact of HVT on Bi-CMPStereo.} Cross-dataset generalization results on the MVSEC~\cite{zhu2018multivehicle} dataset.}
\label{tab5}
\vspace{-3.5em}
\end{table}

\section{Conclusion}

We introduced Bi-CMPStereo, a novel bidirectional framework for event-frame asymmetric stereo that preserves domain-specific features through cross-modal prompting. Our CMPStereo network, equipped with Stereo Canonicalization Constraint and Cross-Domain Embedding Adapter, learns high-fidelity aligned representations in target canonical spaces. Applying Hierarchical Visual Transformation in context learning further boosts the model’s robustness and generalization. By combining complementary features from both event and image domains, our approach effectively addresses the modality gap and achieves robust disparity estimation that surpasses state-of-the-art methods. Extensive experiments demonstrate strong performance and generalization on DSEC, MVSEC and M3ED datasets.

\twocolumn[{
\renewcommand\twocolumn[1][]{#1}
{
    \clearpage
    \centering
    \Large
    \textbf{\thetitle}\par
    \vspace{1.0em}
    Supplementary Material\par
    \vspace{1.0em}
}
}]

This supplementary document provides additional information for the paper "Bidirectional Cross-Modal Prompting for Event-Frame Asymmetric Stereo", including an extended description of the event representations used in our networks (sec.~\ref{sec:supp:event}), details of the train–test split of the DSEC~\cite{gehrig2021dsec} dataset (sec.~\ref{sec:supp:dsec}), and extensive qualitative visual comparisons on DSEC (sec.~\ref{sec:supp:result}).

\section{Event Representation}
\label{sec:supp:event}

Different from conventional cameras that capture intensity frames at a fixed rate, event cameras asynchronously report the per-pixel illumination variation (i.e., event) with microsecond resolution. Letting $L\left( {x,y,t} \right)$ define the logarithmic brightness on pixel $\left( {x,y} \right)$ at time $t$, an event ${e=\left( {x,y,t,p} \right)}$ is generated whenever the log-intensity change within the pixel $\left( {{x},{y}} \right)$ reaches a threshold ${C_{th}}$:
\begin{equation}
	L\left( {{x},{y},{t}} \right) - L\left( {{x},{y},{t} - \Delta t} \right) = {p} \cdot {C_{th}}
\end{equation}
where ${\Delta t}$ is the time elapsed since the previous event at the same pixel, and the polarity ${p} \in \left\{ {+1,-1} \right\}$ indicates whether the brightness has increased or decreased.

Owing to the asynchronous nature of events, we need to convert the event stream into a tensor-like representation that can be interpreted by neural networks. 
To better leverage the complementary cues from image and event domains in our bidirectional framework, imgCMPStereo and evCMPStereo adopt event representations tailored to their respective target domains.

\textbf{Voxel grid \cite{zhu2019unsupervised} for imgCMPStereo.}
Voxel grid takes a spatial-temporal quantization approach. Given a set of $N$ events $\left\{ {{e_i}} \right\}_{i = 1}^N$, by discretizing its duration $\Delta T = {t_N} - {t_1}$ into $B=5$ uniform bins, each event ${e_i}=\left( {{x_i},{y_i},{t_i},{p_i}} \right)$ distributes its polarity to the two nearest voxels using bilinear sampling kernel as follows:
\begin{equation}
	V\left( {x,y,t} \right) = \sum\limits_{{x_i} = x{\rm{ }}{y_i} = y} {{p_i}\cdot\max \left( {0,1 - \left| {t - \widetilde {{t_i}}} \right|} \right)}
\end{equation}
where $\widetilde {{t_i}}: = \frac{{B - 1}}{{\Delta T}}\left( {{t_i} - {t_1}} \right)$ is the normalized timestamp.

Given the proven effectiveness of voxel grid representation in event-based vision tasks, we adopt it in imgCMPStereo to enable efficient cross-modal alignment and stereo matching within the image domain.

\textbf{Event concentration \cite{nam2022stereo} for evCMPStereo.}
A stream of events is initially represented using a mixed-density stacking method, where events are reversely stacked from the depth timestamp with exponentially increasing batch sizes. The sequence consists of $M \!\!=\!\! 10$ stacks and each generates a one-channel tensor ${S_{1 \ldots M}} \in \mathbb{R}^{H \times W}$. The mixed-density event tensors are concatenated along the channel dimension and fed into an attention-based network for weighting the importance of each event stack. A weighted sum is then performed with the output weights ${W} \in \mathbb{R}^{H \times W \times M}$ over the stacked tensors to obtain the concentrated event representation $E \in \mathbb{R}^{H \times W}$:
\begin{equation}
	E\left( {x,y} \right) = \sum\limits_{j = 1}^M {W\left( {x,y,j} \right) \cdot } {S_j}\left( {x,y} \right)
\end{equation}

By concentrating events into a sharp, blur-free edge-like tensor that preserves the intrinsic response characteristics of event cameras, event concentration provides strong complementarity to the image domain and is therefore adopted as the event representation for evCMPStereo.

\section{Details of the DSEC Train–Test Split}
\label{sec:supp:dsec}

We follow \cite{bartolomei2025depth} to split the DSEC~\cite{gehrig2021dsec} dataset, which contains 41 sequences in total, into 31 training sequences and 10 testing sequences. The test set comprises 7,109 samples and includes the following sequences: \textit{zurich\_city\_05\_a}, \textit{zurich\_city\_05\_b},  \textit{zurich\_city\_06\_a}, \textit{zurich\_city\_07\_a}, \textit{zurich\_city\_08\_a}, \textit{zurich\_city\_09\_d}, \textit{zurich\_city\_10\_b}, \textit{interlaken\_00\_f}, \textit{interlaken\_00\_g}, and \textit{thun\_00\_a}. The remaining 31 sequences with 18,950 samples are used for training.

\section{Additional Qualitative Results on DSEC}
\label{sec:supp:result}

\begin{figure*}
    \centering
    \includegraphics[width=0.95\textwidth, page=1, trim=5 5 5 5, clip]{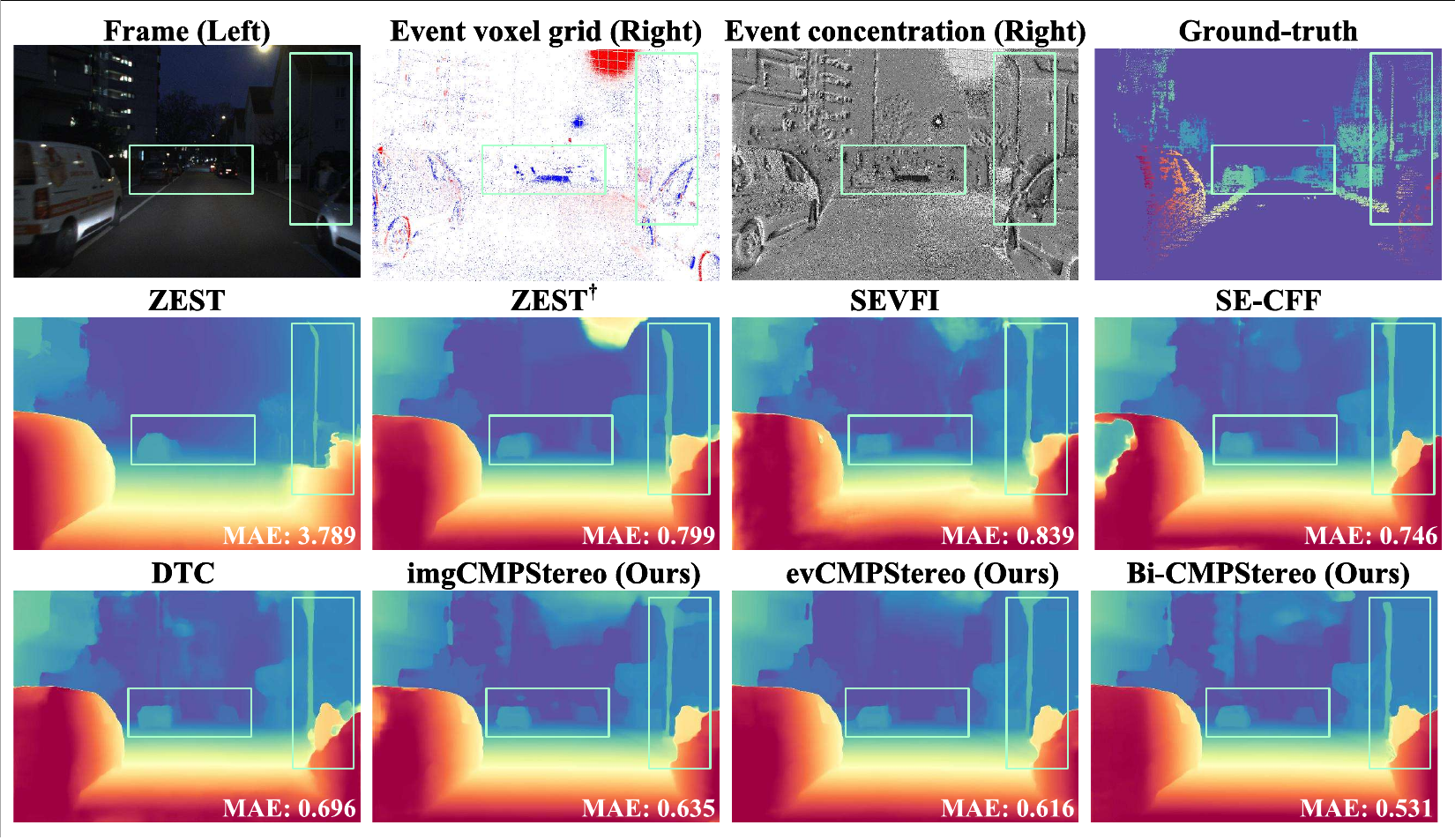}
    \vspace{-0.5em}
    \caption{\textbf{Additional qualitative results in nighttime scenario from the DSEC dataset \cite{gehrig2021dsec}.} The mean absolute error (MAE) is shown at the bottom right of each estimated disparity map.}
    \label{supp:fig1}
    \vspace{-4em}
\end{figure*}

\begin{figure*}
    \centering
    \includegraphics[width=0.95\textwidth, page=2, trim=5 5 5 5, clip]{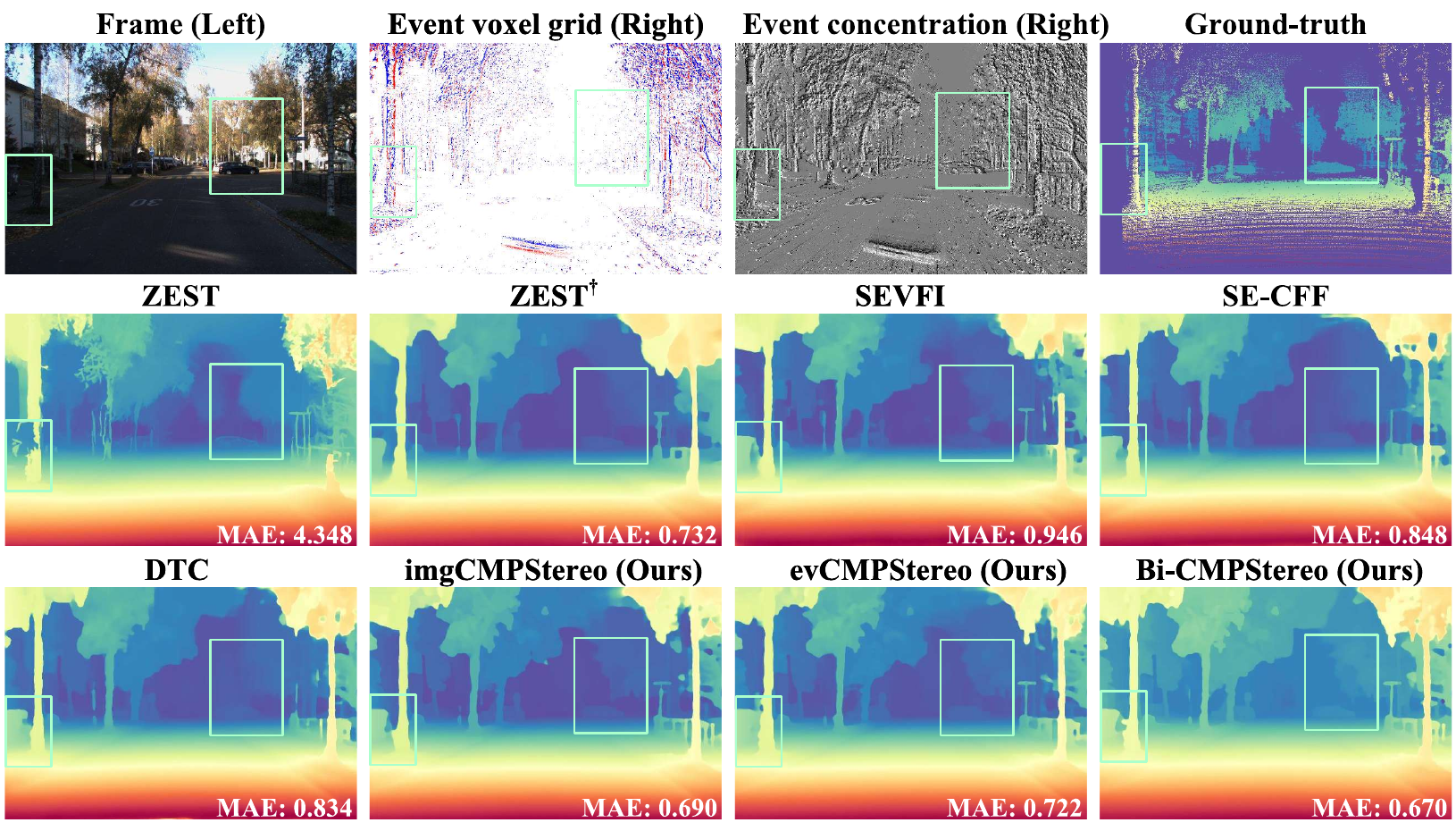}
    \vspace{-0.5em}
    \caption{\textbf{Additional qualitative results in daytime scenario from the DSEC dataset \cite{gehrig2021dsec}.}}
    \label{supp:fig2}
    \vspace{-4em}
\end{figure*}

We provide additional qualitative results on the DSEC \cite{gehrig2021dsec} dataset. As in the main paper, we compare our imgCMPStereo, evCMPStereo, and Bi-CMPStereo with the state-of-the-art event–frame asymmetric stereo method ZEST \cite{lou2024zero}, its DSEC-trained variant ZEST$^{\dag}$, and SEVFI-Net \cite{ding2024video}, as well as two state-of-the-art event-based symmetric stereo approaches, SE-CFF \cite{nam2022stereo} and DTC-SPADE \cite{zhang2022discrete}.
Fig.~\ref{supp:fig1} and ~\ref{supp:fig2} are results in nighttime and daytime scenarios, respectively. We additionally present the event concentration, which clearly shows the scene edges and demonstrates it preservation of the edge–response characteristics of the event camera. The results highlight the advantages of our approach in producing higher accuracy and higher-quality structural disparity maps. In the highlighted boxes, despite the ambiguity introduced by complex textures or low-light conditions, our method still constructs sharp edges and fine structural details.



\textbf{Acknowledgement.} This work was supported by Major Science and Technology Major Project of Jiangsu Province (BG2025025).
{
	\small
	\bibliographystyle{ieeenat_fullname}
	\bibliography{main}
}

\end{document}